\documentclass[letterpaper, 10pt, journal]{IEEEtran}  

\usepackage{mathtools,amsfonts,amssymb,dsfont,xcolor}
\usepackage{amsthm,amsmath}
\usepackage{thm-restate}
\usepackage[sort,compress]{cite}

\DeclareMathOperator{\VEC}{vec}
\DeclareMathOperator{\TR}{Tr}
\DeclareMathOperator{\diag}{diag}

\newcommand\OO{\mathcal O}
\newcommand\T{\mathcal{T}}

\newcommand\Ind{\mathds{1}}

\newcommand\reals{\mathds{R}}
\newcommand\integers{\mathds{N}}
\newcommand\EXP{\mathds{E}}
\newcommand\PR{\mathds{P}}

\DeclareMathOperator*{\argmin}{arg\,min}

\newtheorem{assumption}{Assumption}

\newtheorem{definition}{Definition}

\newtheorem{theorem}{Theorem}

\newtheorem{corollary}{Corollary}
\newtheorem{lemma}{Lemma}
\newtheorem{proposition}{Proposition}
\theoremstyle{definition}
\newtheorem{example}{Example}
\newtheorem{remark}{Remark}

\newcommand*\TRANS{{\mathpalette\doTRANS\empty}}
\makeatletter
\newcommand*\doTRANS[2]{\raisebox{\depth}{$\m@th#1\intercal$}}
\makeatother

\title{Strong Consistency and Rate of Convergence of Switched Least Squares System Identification for Autonomous Markov Jump Linear Systems}

\author{Borna Sayedana, Mohammad Afshari, Peter E. Caines, Aditya Mahajan%
\thanks{\noindent An earlier version of this paper is presented in 61st Conference on Decision and Control (CDC), Cancun, Mexico, 2022~\cite{sayedana2021consistency}.}%
\thanks{The authors are with the Department of Electrical and Computer Engineering, McGill University, 3480 Rue University, Montreal, QC H3A\,0E9, Canada. Emails: \texttt{\{borna.sayedana, mohammad.afshari2\}@mail.mcgill.ca, \{peter.caines, aditya.mahajan\}@mcgill.ca}.}%
\thanks{The work was supported in part by Fonds de Recherche
du Québec, Nature et Technologies (FRQNT), Bourses de doctorat en recherche Grant 316558 and  
in part by Natural Sciences and Engineering
Research Council of Canada (NSERC) Discovery Grant
RGPIN-2021-03511.}}

\begin{document}

\maketitle

\begin{abstract}%
In this paper, we investigate the problem of system identification for autonomous Markov jump linear systems (MJS) with complete state observations.
We propose switched least squares method for  identification of MJS, show that this method is strongly consistent, and derive  data-dependent and data-independent rates of convergence.
In particular, our data-independent rate of convergence shows that, almost surely, the system identification error is $\OO\big(\sqrt{\log(T)/T} \big)$  where $T$ is the time horizon. These results show that switched least squares method for MJS has the same rate of convergence as least squares method for autonomous linear systems.
We derive our results by imposing a general stability assumption on the model called stability in the average sense. We show that stability in the average sense is a weaker form of stability compared to the stability assumptions commonly imposed in the literature. We present numerical examples to illustrate the performance of the proposed method.
\end{abstract}


\section{Introduction}

 Markov jump linear systems (MJS) are a good approximation of non-linear time-varying systems arising in various applications including networked control systems \cite{deaecto2014discrete} and  cyber-physical systems \cite{de2015input,cetinkaya2018analysis}. There is a rich literature on the stability analysis (e.g., \cite{fang1994almost, fang1997new,costa2006discrete}) and optimal control (e.g., \cite{chizeck1986discrete}) of MJS. However, most of the literature assumes that the system model is known. The question of system identification, i.e., identifying the dynamics from data, has not received much attention in this setup.

 The problem of identifying the system model from data is a key component for control synthesis for both offline control methods and online control methods including adaptive control and reinforcement learning \cite{goodwin1980discrete,ljung1998system}. There are four main approaches for system identification of linear systems: (i)~maximum likelihood estimation  which maximizes the likelihood function of the unknown parameter given the observation (e.g. see \cite{rissanen1979strong}); (ii)~minimum prediction error methods which minimize the estimation error (residual process) according to some loss function (e.g. see \cite{ljung1976consistency,caines1976prediction}); (iii)~subspace methods, which find a minimum state space realization given the input, output data (e.g. see \cite{ho1966effective,10.1007/978-94-009-8546-9_10}); (iv)~least squares method which estimates the unknown parameter by considering the model as a regression problem (e.g. see \cite{lai1985asymptotic,caines2018linear}).
 
 These methods differ in terms of structural assumptions on the model (e.g. system order), hypotheses on the stochastic process, and convergence properties and guarantees.
 
 Structural assumptions require the system to be stable in some sense (e.g., mean square stable, exponentially stable, etc.), and stochastic hypotheses restrict the noise processes to be  of a certain type, (e.g., Gaussian, sub-Gaussian, or Martingale difference sequences). 
 	
Convergence properties characterize the asymptotic behavior of system identification methods. The basic requirements for any system identification method is its consistency, asymptotic normality and rates of convergence, that is to establish that estimates converge asymptotically to the true unknown parameter and characterize the rate of convergence. System identification methods can be \emph{weakly}  consistent (i.e., estimates converge in probability) or \emph{strongly} consistent (i.e., estimates convergence almost surely). For linear systems, there is a vast literature that establishes the consistency and rates of convergence for a variety of methods (e.g. see \cite{caines2018linear,ljung1998system} for a unified overview). Another characterization of the convergence is finite-time guarantees which provide lower-bounds on the number of samples required so that estimates have a specified degree of accuracy with a specified high probability~\cite{faradonbeh2020optimism,faradonbeh2018finite, abbasi2011regret,faradonbeh2020input,simchowitz2018learning,oymak2019non,zheng2021sample,lale2020logarithmic,tsiamis2022statistical}. As the number of samples grow to infinity, these results establish weak consistency of the proposed methods.

  System identification of MJS and switched linear systems (SLS) has received less attention in the literature. There is some work on designing asymptotically stable controllers for unknown SLS \cite{caines1985optimal,caines1995adaptive,xue2001necessary} but these papers do not establish rates of convergence for system identification.
 There are some recent papers which provide finite time guarantees and rate of convergence for SLS~\cite{9810989,sarkar2019nonparametric} and MJS~\cite{sattar2021identification}. System identification of a globally asymptotically stable SLS with controlled switching signal is investigated in \cite{9810989}, while the system identification of an unknown order SLS using subspace methods is investigated in  \cite{sarkar2019nonparametric}. Both these methods are developed for SLS and are not directly applicable to MJS. The model analyzed in \cite{sattar2021identification} is an MJS system. Under the assumption that the system is mean square stable, the switching distribution is ergodic and the noise is i.i.d.\ subgaussian, it is established that the convergence rate is $\OO(\sqrt{\log T/T})$ with high probability. Then a certainty equivalence control algorithm is proposed and its regret is analyzed. Note that if we let the number of samples go to infinity, these results imply \emph{weak} consistency of the proposed methods for MJS systems. As far as we are aware, there is no existing result which establishes \emph{strong} consistency of a method for system identification of MJS.

\subsection{Contributions}
\begin{itemize}
	\item We propose a \emph{switched} least squares method for system identification of an unknown (autonomous) MJS  and provide data-dependent and data-independent rates of convergence for this method.
	\item  We prove \emph{strong} consistency of the switched least squares method and establish a  $\OO(\sqrt{\log(T)/T})$ rate of convergence, which matches with the rate of convergence of non-switched linear systems established in \cite{lai1985asymptotic}. In contrast to the existing high-probability convergence guarantees in the literature, our results show that the estimates converge to the true parameters \emph{almost surely}. Therefore, our results provide guarantees which are different in nature compared to parallel works.
	\item The main challenge in establishing strong consistency for MJS systems is the interplay between the empirical covariance process and stability of the MJS system. We shed light on this connection and show that stability in the average sense is a sufficient condition for strong consistency.
	\item Our results are derived under weaker assumptions compared to the existing literature (Note that the preliminary version of this paper \cite{sayedana2021consistency} had assumed a stronger stability condition). Most existing results assume that the MJS system is mean square stable. We prove that mean square stability implies stability in the average sense. Furthermore, we show that a commonly used sufficient condition for almost sure stability of noise-free MJS system also implies stability in the average sense.  
\end{itemize}


\subsection{Organization}The rest of the paper is organized as follows. In Sec~\ref{sec:problem_formulation}, we present the system model, assumptions, and the main results. In Sec.~\ref{sec:proof_main_1}, we prove the main results. In Sec.~\ref{sec:stab_avg}, we explain the connection of stability in the average sense with mean square stability and almost sure stability.  We present an illustrative example in Sec.~\ref{sec:numerical_simulation}. We conclude in Sec.~\ref{sec:conclusion}.

\subsection{Notation}\label{sec:notation}
Given a matrix $A$, $A(i,j)$ denotes its $(i,j)$-th element, $\lambda_{\max}(A)$ and $\lambda_{\min}(A)$ denote the largest and smallest magnitudes of right eigenvalues, $\sigma_{\max}(A) = \sqrt{\lambda_{\max}(A^{\TRANS} A)}$ denotes the spectral norm. For a square matrix $Q$, $\TR(Q)$ denotes the trace. When $Q$ is symmetric, $Q \succeq 0$ and $Q \succ 0$ denotes that $Q$ is positive semi-definite and positive definite, respectively. For two square matrices, $Q_{1}$ and $Q_{2}$ of the same dimension, $Q_{1} \succeq Q_{2}$ means $Q_{1}-Q_{2} \succeq 0$. Given two matrices $A$ and $B$, $A \otimes B$ denotes the Kronocher product of the two matrices.

Given a sequence of positive numbers $\{a_{t}\}_{t \geq 0},a_{T} = \OO(T)$ means that  $\limsup_{T \rightarrow \infty}a_{T}/T < \infty$, and $a_{T} = o(T)$ means that  $\limsup_{T \rightarrow \infty}a_{T}/T = 0$.
Given a sequence of vectors $\{ x_t \}_{t \in \mathcal{T}}$, $\VEC(x_{t})_{t \in \T}$ denotes the vector formed by vertically stacking $\{x_{t}\}_{t \in \T}$.
Given a sequence of random variables $\{x_{t}\}_{t \geq 0}$, $x_{0:t}$ is a short hand for $(x_{0},\cdots,x_{t})$ and $\sigma(x_{0:t})$ denotes the sigma field generated by random variables $x_{0:t}$. Given a probability space $\{\Omega,\mathcal{F},\PR\}$, $\Omega$ denotes the sample space, $\omega \in \Omega$ denotes elementary events, $\PR(\cdot)$ denotes the probability measure and $\EXP[\cdot]$ denotes the expectation operator.

$\reals$ and $\integers$ denote the sets of real and natural numbers. For a set $\mathcal{T}$, $|\mathcal{T}|$ denotes its cardinality. For a vector $x$, $\| x \|$ denotes the Euclidean norm. For a matrix $A$, $\| A \|$ denotes the spectral norm and $\| A \|_{\infty}$ denotes the element with the largest absolute value. $\diag(\cdot)$ is the block diagonal matrix. Convergence in almost sure sense is abbreviated as  $a.s$.

\section{System model and problem formulation}\label{sec:problem_formulation}

Consider a discrete-time (autonomous) MJS. The state of the
system has two components: a discrete component $s_t \in \mathcal{S} \coloneqq \{1, \dots, k\}$ and
a continuous component $x_t \in \reals^n$. There is a finite set $\mathcal{A}
= \{A_1, \dots, A_k\}$ of system matrices, where $A_i \in \reals^{n\times n}$.
The continuous component~$x_t$ of the state starts at a fixed value $x_0$ and the initial discrete state $s_{0}$ starts according to a prior distribution $\pi_{0}$. The continuous state evolves according to:
\begin{equation}\label{eq:model}
  x_{t+1} = A_{s_t} x_t + w_{t}, 
  \quad t \ge 0,
\end{equation}
where $\{w_t\}_{t \ge 0}$, $w_t \in \reals^n$, is a noise process. The discrete component evolves in a Markovian manner according to a time-homogeneous irreducible and aperiodic transition matrix $P$, i.e. $\PR(s_{t+1} = j|s_{t} = i) = P_{ij}$. 

Let $\pi_{t} = (\pi_t(1),\ldots,\pi_{t}(k)) $ denote the probability distribution of the discrete state at time $t$ and $\pi_{\infty}$ denote the stationary distribution. We assume $\pi_{\infty}(i) \not=0$ for all $i$. Let $\mathcal{F}_{t-1} = \sigma(x_{0:t}, s_{0:t})$ denote the sigma-algebra
generated by the history of the complete state.  

It is assumed that the noise process satisfies the following:
\begin{assumption}\label{ass:noise}
	The noise process $\{w_t\}_{t \ge 0}$ is a martingale difference sequence
	with respect to $\{\mathcal{F}_t\}_{t \ge 0}$, i.e., $\EXP[ |w_t| ] < \infty$
	and $\EXP[ w_t \mid \mathcal{F}_{t-1} ] = 0$. Furthermore, there exists a
	constant $\alpha > 2$ such that $
 \sup_{t \ge 0} \EXP[ \| w_t \|^\alpha \mid \mathcal{F}_{t-1} ] < \infty
	 \quad \text{a.s.}$
	and there exists a symmetric and positive definite matrix $C \in \reals^{n
		\times n}$ such that
  $\liminf_{T \to \infty} \frac{1}{T} \sum_{t=0}^{T-1} w_{t} w_{t}^\TRANS 
	= C \quad \text{a.s.}$
\end{assumption}
Assumption~\ref{ass:noise} is a standard assumption in the asymptotic analysis of system identification of linear systems~\cite{caines2018linear,lai1982least,lai1985asymptotic,chen1986convergence,chen1987optimal} and allows the noise process to be non-stationary and have heavy tails (as long as moment condition is satisfied). We use the following notion of stability for the MJS system~\eqref{eq:model}.
\begin{definition}\label{def:avg}
	The MJS system~\eqref{eq:model} is called stable in the \emph{average sense} if almost surely:
	\[
	\sum_{t=1}^{T}\|x_{t}\|^2 = \OO(T) \quad \text{i.e.} \quad
	\limsup_{T \rightarrow \infty}\frac{1}{T}\sum_{t=1}^{T}\|x_{t} \|^2 < \infty.
	\]
\end{definition}
	\begin{assumption}\label{ass:stab_avg}
		The MJS system~\eqref{eq:model} is stable in the average sense.
	\end{assumption}
The notion of stability in the average sense has been used in a few papers in the literature of linear systems \cite{duncan1990adaptive},\cite{faradonbeh2020adaptive}; however, in the MJS literature, the commonly used notions of stability are  mean square stability and almost sure stability of \textit{noise-free} system. We compare stability in the average sense with both of these notions in Sec.~\ref{sec:stab_avg}. Specifically, we show that mean square stability implies stability in the average sense. Moreover, we show a common sufficient condition for almost sure stability of noise-free system implies stability in the average sense for MJS system~\eqref{eq:model}.  Therefore, the assumption of stability in the average sense is weaker than the commonly imposed stability assumptions imposed in the literature.

\subsection{System identification and switched least squares estimates} 
We are interested in the setting where the system dynamics
$\mathcal{A}$ and the switching transition matrix  $P$ are unknown. Let
$\theta^\TRANS = [ A_1, \dots, A_k]$ $ \in \reals^{n \times nk}$ denote the
unknown parameters of the system dynamics matrices. We consider an agent
that observes the complete state $(x_{t}, s_{t})$ of the system at each time and
generates an estimate $\hat \theta_{T}$ of $\theta$ as a function of the observation history
$(x_{0:T}, s_{0:T})$. A commonly used estimate in such settings is the least
squares estimate: 
\begin{equation}\label{eq:LS_gen}
	\hat{\theta}_{T}^{\TRANS} = \argmin_{\theta^\TRANS = [ A_1, \dots, A_k]}\sum_{t=0}^{T-1}\|x_{t+1} - A_{s_{t}}x_{t}\|^2.
\end{equation}

The components $[\hat A_{1,T}, \dots, \hat A_{k,T}] = \hat \theta_{T}^\TRANS$ of the least squares estimate can be computed in a switched manner.  Let $\mathcal{T}_{i,T} = \{ t \le T \mid s_{t} = i \}$ denote the time indices until time~$T$ when the discrete state of the system equals~$i$. Note that for each $t \in \T_{i,T}$, $A_{s_{t}} = A_{i}$. Therefore, we have
\begin{equation}\label{eq:LS_Problem}
  \hat{A}_{i,T} \coloneqq \argmin_{A_{i} \in \reals^{n \times n} }\sum_{t \in \mathcal{T}_{i,T}}\|x_{t+1} - A_{i}x_{t}\|^2 , \quad \forall i \in \{1,\cdots,k\}.
\end{equation}
Let $X_{i,T}$ denote  $\sum_{t \in \T_{i,T}}^{}x_{t}x_{t}^{\TRANS}$, which we call the unnormalized empirical covariance of the continuous component of the state at time $T$ when the discrete component equals $i$. Then, $\hat{A}_{i,T}$ can be computed recursively as follows:
\begin{multline}
	\hat{A}_{i,T+1} = \hat{A}_{i,T}  \\+  \bigg[ \frac{X_{i,T}^{-1}x_{T}(x_{T+1} - \hat{A}_{i,T}x_{T})^{\TRANS}}{1+x_{T}^{\TRANS}X_{i,T}^{-1}x_{T}}\bigg] \Ind\{s_{T+1} = i\}
\end{multline}
where $X_{i,T}$ may be updated as
$
	X_{i,T+1} = X_{i,T} +\bigl[ x_{T+1}x^{\TRANS}_{T+1}\bigr]\Ind\{s_{T+1} = i\}.
$
Due to the switched nature of the least squares estimate, we refer to above estimation procedure as \emph{switched least squares} system identification.  
%
%
%

\subsection{The main results}
A fundamental property of any sequential parameter estimation method is strong consistency, which we define below.

\begin{definition}
	An estimator $\hat{\theta}_{T}$ of parameter $\theta$ is called strongly consistent if $
	\lim_{T \rightarrow \infty} \hat{\theta}_{T} = \theta$, a.s.
\end{definition}
Our main result is to establish that the switched least squares estimator is strongly consistent. We do so by providing two different characterization of the rate of convergence. We first provide a data-dependent rate of convergence which depends on the spectral properties of the unnormalized empirical covariance. We then present a data-independent characterization of rate of convergence which only depends on $T$. All the proofs are presented in Sec.~\ref{sec:proof_main_1}.

\begin{theorem}\label{thm:main_1}
  Under Assumptions~\ref{ass:noise} and~\ref{ass:stab_avg}, the switched least squares estimates $\{\hat A_{i,T}\}_{i = 1}^k$ are strongly consistent, i.e., for each $i \in \mathcal{S}$, we have:
    $\lim_{T \to \infty} \bigl\| \hat A_{i,T} - A_i \bigr\|_\infty = 0,$ a.s.
   Furthermore, the rate of convergence is upper bounded by:
  \[
    \bigl\| \hat A_{i,T} - A_i \bigr\|_\infty \le
    \OO\bigg(\sqrt{\frac{\log\big[\lambda_{\max}(X_{i,T})\big]}{\lambda_{\min}(X_{i,T})}}\bigg), \quad \text{a.s.}
  \] 
\end{theorem}
\begin{remark}\label{rem:covraiates}
Theorem~\ref{thm:main_1} is not a direct consequence of the decoupling procedure in the switched least squares method. The $k$ least squares problems have a common covariate process $\{x_{t}\}_{t \geq 1}$. Therefore, the convergence of the switched least squares method and the stability of the MJS are interconnected problems. Our proof techniques carefully use the stability properties of the system to establish the consistency of the system identification method.
\end{remark}
We  simplify the result of Theorem~\ref{thm:main_1} and characterize the data dependent result of Theorem~\ref{thm:main_1} in terms of horizon $T$ and the cardinality of the set $\mathcal{T}_{i,T}$.
\begin{corollary}\label{cor:2}
	Under Assumptions~\ref{ass:noise} and~\ref{ass:stab_avg},   for each $i \in \mathcal{S}$, we have:
	  \[
	\bigl\| \hat A_{i,T} - A_i \bigr\|_\infty \le
	\OO\bigg(\sqrt{\log(T)/|\T_{i,T}|}\bigg), \quad \text{a.s.}
	\]
\end{corollary}
\begin{remark}
	The assumption that $\pi_{\infty}(i) \not = 0$ implies that for sufficiently large $T$,  $|\mathcal{T}_{i,T}| \not= 0$ almost surely, therefore the expressions in above bounds are well defined.  
\end{remark}
The result of Corollary~\ref{cor:2} still depends on  data. When system identification results are used for adaptive control or reinforcement learning, it is useful to have a data-independent characterization of the rate of convergence. We present this characterization in the next theorem.


\begin{theorem}\label{thm:main_2}
	Under Assumptions~\ref{ass:noise} and~\ref{ass:stab_avg}, the rate of convergence of the switched least squares estimator $\hat{A}_{i,T}$, $i \in \mathcal{S}$ is upper bounded by:
	\[
	\bigl\| \hat A_{i,T} - A_i \bigr\|_\infty \le
	\OO\big(\sqrt{\log(T)/\pi_{\infty}(i)T}\big), \quad \text{a.s.}
	\]
	where the constants in the $\OO(\cdot)$ notation do not depend on Markov chain $\{s_{t}\}_{t\geq 0}$ and horizon $T$. Therefore, the estimation process $\{\hat{\theta}_{T}\}_{T\geq1}$ is strongly consistent, i.e.,
	$\lim_{T \to \infty} \bigl\| \hat \theta_{T} - \theta \bigr\|_\infty = 0 $ a.s.\ Furthermore, the rate of convergence is upper bounded by:
	\begin{equation*}\label{eq:error_rates}
	\bigl\| \hat \theta_{T} - \theta \bigr\|_\infty \le
	\OO\big(\sqrt{\log(T)/\pi^{*}T}\big), \quad \text{a.s.}
	\end{equation*}
	where $\pi^{*} = \min_{j \in \mathcal{S}} \pi_{\infty}(j) $.
\end{theorem}
 Theorem~\ref{thm:main_2} shows that  Assumptions~\ref{ass:noise} and \ref{ass:stab_avg} guarantee that the switched  least squares estimator for MJS has the same rate of convergence of $\OO(\sqrt{\log(T)/T})$ as non-switched case  established in \cite{lai1985asymptotic}. Moreover, the upper bound in Theorem~\ref{thm:main_2} shows that the estimation error of $\hat{A}_{i,T}$ is proportional to $1/\sqrt{\pi_{\infty}(i)}$; therefore, the rate of convergence of $\hat{\theta}_{T}$ is proportional to $1/\sqrt{\pi^{*}}$, where $\pi^{*}$ is the smallest probability in the stationary distribution $\pi_\infty$.
 
 \begin{remark}
 	SLS is a special case of MJS in which the discrete state evolves in an i.i.d. manner. The results presented in this section are valid for the SLS after substituting stationary distribution $\pi_{\infty}$ with the i.i.d. PMF of switching probabilities defined over discrete state. 
 \end{remark}

%
\section{Proofs of the main results}\label{sec:proof_main_1}

\subsection{Preliminary results}
We first state the Strong Law of Large Numbers (SLLN) for Martingale Difference Sequences (MDS).
\begin{theorem}(see \cite[Theorem 3.3.1]{stout1974almost})\label{thm:SLLN_MDS}
	Suppose $\{X_{\tau}\}_{\tau \geq 1}$is a martingale difference sequence with respect to the filtration $\{\mathcal{F}_{\tau}\}_{ \tau \geq 1}$ . Let $a_{\tau}$ be $\mathcal{F}_{\tau-1}$ measurable for each $\tau \geq 1$ and we have $0<a_{\tau} \rightarrow \infty$ as $\tau \rightarrow \infty$, a.s. If for some $p \in (0,2]$, we have:
	$
	\sum_{\tau=1}^{\infty}\EXP [ |X_{\tau}|^{p} | \mathcal{F}_{\tau-1}]/a_{\tau}^{p} < \infty,
	$
	then:
	$
	\lim_{T \to \infty} \sum_{\tau=1}^{T}X_{\tau}/a_{T} = 0 \quad \text{a.s.}
	$
\end{theorem}

\begin{lemma}\label{lem:occ_measure}
	The assumptions on the process $\{s_{t}\}_{t \geq 0}$ imply that $\lim_{T \to \infty} |\T_{i,T}|/T = \pi_{\infty}(i)$, a.s.
\end{lemma}
\begin{proof}
     $\{ s_t \}_ {t \geq 0}$  is an aperiodic and irreducible Markov chain,  hence, by the Ergodic Theorem (Theorem 4.1, \cite{bremaud2013markov}),   $\{ s_t \}_ {t \geq 0}$  is ergodic and  therefore $\lim_{T \to \infty} |\T_{i,T}|/T = \pi_{\infty}(i)$ a.s.
\end{proof}
\begin{lemma}\label{lem:local_growth}
	Assumption~\ref{ass:noise} and \ref{ass:stab_avg} imply:
	\[
	\sum_{\tau=1}^{\infty} \|x_{\tau}\|^2/\tau^2 < \infty \quad \text{a.s.}
	\]
\end{lemma}

\begin{proof}
	The result is a direct consequence of Abel's lemma. 
	Let $S_{T} \coloneqq \sum_{\tau=1}^{T}\|x_{\tau} \|^2$, then we have:
	\begin{align*}
		\sum_{\tau=1}^{T}\frac{\|x_{\tau}\|^2}{\tau^2} &= \sum_{\tau=1}^{T}\frac{S_{\tau} - S_{\tau-1}}{\tau^2} \\ &= \frac{S_{T}}{T^2} - \frac{S_{0}}{1} + \sum_{\tau=2}^{T}S_{\tau-1}\Big(\frac{1}{(\tau-1)^2} - \frac{1}{\tau^2}\Big)\\&\stackrel{(a)}{=}\frac{S_{T}}{T^2} - \frac{S_{0}}{1} + \sum_{\tau=2}^{T}\OO(\tau-1)\Big(\frac{2\tau-1}{\tau^2(\tau-1)^2}\Big) \\&=\frac{S_{T}}{T^2} - \frac{S_{0}}{1} +\sum_{\tau = 2}^{T}\OO\bigg(\frac{1}{\tau^2}\bigg)< \infty
	\end{align*}
	where $(a)$ follows from Assumption~\ref{ass:stab_avg}.
\end{proof} 
\begin{lemma}\label{lem:cross}
	We have the following:
	\[
	\Big\|\sum_{\tau=1}^{T}A_{s_\tau}x_{\tau}w^{\TRANS}_{\tau}  + w_{\tau}x^{\TRANS}_{\tau}A^{\TRANS}_{s_{\tau}}\Big\| = o(T) \quad  \text{a.s.}
	\]
\end{lemma}
\begin{proof}
	We prove the limit element-wise. The $(l,p)$-th element of the matrix $A_{s_{\tau}}x_{\tau}w^{\TRANS}_{\tau}$ is
	$
	\Big[\sum_{j=1}^{n} A_{s_{\tau}}(l,j)x_{\tau}(j)\Big]w_{\tau}(p).
	$
	We calculate the term:
	\begin{equation}
	   \EXP\Big[\Big( \sum_{j=1}^{n} A_{s_{\tau}}(l,j)x_{\tau}(j)w_{\tau}(p)\Big)^2\Big| \mathcal{F}_{\tau-1} \Big].   
	\end{equation}
	%
	Let $A_{*} = \max_{i \in \mathcal{S}} \|A_{i} \|_{\infty}$, then
	\begin{align*}
	   \hskip 2em & \hskip -2em
    \EXP\Big[\Big(\sum_{j=1}^{n} A_{s_{\tau}}(l,j)x_{\tau}(j)\Big)^2w^2_{\tau}(p) \Big| \mathcal{F}_{\tau-1} \Big] \\&\stackrel{(a)}{\leq} A_{*}^2 \sup_{\tau}\EXP[w^2_{\tau}(p)\big|\mathcal{F}_{\tau-1}]\Big(\sum_{j=1}^{n}x_{\tau}(j)\Big)^2 \\
		&\stackrel{(b)}{\leq} nA_{*}^2 \sup_{\tau}\EXP\big[w^2_{\tau}(p)\big|\mathcal{F}_{\tau-1}\big] \|x_{\tau} \|^2,
	\end{align*}
	where $(a)$ uses the fact that $s_{\tau}$ and $x_{\tau}$ are $\mathcal{F}_{\tau-1}$ measurable and that $|A_{s_{\tau}}(l,j)| \leq A_{*}$ and  $(b)$ is by Cauchy-Schwarz's inequality.
	Therefore:
	\begin{align*}
        \hskip 2em & \hskip -2em
		\sum_{\tau=1}^{T} \frac{\EXP\Big[\Big( \big[\sum_{j=1}^{n} A_{s_{\tau}}(l,j)x_{\tau}(j)\big]w_{\tau}(p)\Big)^2\Big| \mathcal{F}_{\tau-1} \Big]}{\tau^2} \\
       &\leq
		nA_{*}^2\sup_{\tau}\Big\{ \EXP[w_{\tau}^2(p)|\mathcal{F}_{\tau -1}]\Big\}\sum_{\tau = 1}^{T} \frac{\|x_{\tau} \|^2}{\tau^2} \stackrel{(c)}{\leq} \infty.
	\end{align*}
	Since $\alpha > 2$ in Assumption~\ref{ass:noise}, and finiteness of higher order moments imply finiteness of lower order moments, we get   $\EXP\big[w^2_{\tau}(p)\big|\mathcal{F}_{\tau-1}\big]$ is uniformly bounded. 
		This fact along with Lemma~\ref{lem:local_growth} imply $(c)$. The result then follows by applying Theorem~\ref{thm:SLLN_MDS} by setting $a_{t} = t$ and $p=2$.
\end{proof}

We characterize the asymptotic behavior of  the matrix $X_{i,T}$. 
\begin{proposition}\label{prop:main}
	Under Assumptions~\ref{ass:noise} and~\ref{ass:stab_avg}, the following hold a.s.\ for each $i \in  \mathcal{S}$:\\
	\textup{(\textbf{P1})} $\lambda_{\max}(X_{i,T}) = \OO(T)$, $a.s.$\\
	\textup{(\textbf{P2})} $\liminf_{T \rightarrow \infty}\lambda_{\min}(X_{i,T})/|\T_{i,T}| > 0$, $a.s$.
\end{proposition}
\begin{remark}
Property (P1) shows that when the system is stable in the average sense, $\lambda_{\max}(X_{i,T})$ cannot grow faster than linearly with time. Therefore, the stability of the system controls the rate at which $X_{i,T}$ can grow. Property (P2) shows that when the noise has a minimum covariance, $\lambda_{\min}(X_{i,T})$ cannot grow slower than linearly with time.
\end{remark}
\begin{proof}[Proof of (P1)]
	 The maximum eigenvalue of a matrix can be upper bounded as follows:
\begin{align*}
	\lambda_{\max}\Big(\sum_{t \in \T_{i,T}}^{}x_{t}x^{\TRANS}_{t}\Big) \stackrel{(a)}{\leq} \text{tr}\Big(\sum_{t \in \T_{i,T}}^{}x_{t}x^{\TRANS}_{t}\Big) = \sum_{t \in \T_{i,T}}^{}\|x_{t} \|^2\\  \leq \sum_{t=1}^{T}\|x_{t} \|^2 = \OO(T)
\end{align*}
where $(a)$ follows from the fact that trace of a matrix is sum of its eigenvalues and all eigenvalues of $x_{t}x^{\TRANS}_{t}$ are non-negative. 
\end{proof}

\begin{proof}[Proof of (P2)]
For $\tau \geq 1$,
we have:
\begin{align*}
	x_{\tau} x^{\TRANS}_{\tau} = &(A_{s_{\tau-1}}x_{\tau-1} + w_{\tau-1})(A_{s_{\tau-1}}x_{\tau-1} + w_{\tau-1})^{\TRANS} \\ =
	&A_{s_{\tau-1}}x_{\tau-1}x^{\TRANS}_{\tau-1}A^{\TRANS}_{s_{\tau-1}}  \\+ &A_{s_{\tau-1}}x_{\tau-1}w^{\TRANS}_{\tau-1}  +w_{\tau-1}x^{\TRANS}_{\tau-1}A^{\TRANS}_{s_{\tau-1}} + w_{\tau-1}w^{\TRANS}_{\tau-1}.
\end{align*}
Since $A_{s_{\tau-1}}x_{\tau-1}x^{\TRANS}_{\tau-1}A^{\TRANS}_{s_{\tau-1}} $ is positive semi-definite, we have:		 
\begin{equation*}	 
	x_{\tau}x^{\TRANS}_{\tau} \succeq A_{s_{\tau-1}}x_{\tau-1}w^{\TRANS}_{\tau-1} + w_{\tau-1}x^{\TRANS}_{\tau-1}A^{\TRANS}_{s_{\tau-1}} + w_{\tau-1}w^{\TRANS}_{\tau-1}.
\end{equation*}
By summing over $\tau \in \T_{i,T}$, we get:
\begin{multline*}
	\sum_{\tau \in \T_{i,T}}^{}x_{\tau}x^{\TRANS}_{\tau} \succeq \sum_{\tau \in \T_{i,T} }^{}w_{\tau-1}w^{\TRANS}_{\tau-1} + x_{0}x_{0}^{\TRANS}\Ind\{s_{0} = i\} \\+ \sum_{\tau \in \T_{i,T} }^{} \big[A_{s_{\tau-1}}x_{\tau-1}w^{\TRANS}_{\tau-1} + w_{\tau-1}x^{\TRANS}_{\tau-1}A^{\TRANS}_{s_{\tau-1}}\big]\\
	\stackrel{(a)}{\succeq} \sum_{\tau \in\T_{i,T}}^{}w_{\tau-1}w^{\TRANS}_{\tau-1}  + o(T) \quad \text{a.s.}
\end{multline*}
where $(a)$ follows from Lemma~\ref{lem:cross} and $x_{0}x_{0}^{\TRANS}\Ind\{s_{0} = i\} \succeq 0$. Furthermore, since $\lim_{T \rightarrow \infty} |\T_{i,T}|/T = \pi_{\infty}(i)$ a.s. by Lemma~\ref{lem:occ_measure} and $\pi_{\infty}(i)\not=0$ by assumptions on $\{s_{\tau}\}_{\tau \geq 0}$, we have:
\begin{multline*}
	\liminf_{|\T_{i,T}| \rightarrow \infty} \frac{\sum_{\tau \in \T_{i,T}}^{}x_{\tau}x^{\TRANS}_{\tau}}{|\T_{i,T}|} \\\succeq \liminf_{|\T_{i,T}| \rightarrow \infty}\frac{\sum_{\tau \in \T_{i,T}}w_{\tau-1}w^{\TRANS}_{\tau-1}}{|\T_{i,T}|} \stackrel{(b)}{=} C \succ 0 \quad \text{a.s.}
\end{multline*}
where $(b)$ holds by Assumption~\ref{ass:noise} and independence of $\{w_{\tau}\}_{\tau \geq 0}$ and $\{s_{\tau}\}_{\tau \geq 0}$ processes. Therefore
\[
	\liminf_{|\T_{i,T}| \rightarrow \infty}\lambda_{\min} \Big(\frac{\sum_{\tau \in \T_{i,T}}^{}x_{\tau}x^{\TRANS}_{\tau}}{|\T_{i,T}|}\Big) \succ 0.
\]
\end{proof}
\vspace{-8mm}
\subsection{Background on least square estimator}\label{subsec:prelim}
Given a filtration $\{\mathcal{G}_{t}\}_{t \geq 0}$, consider the following regression model:
\begin{equation}\label{eq:Lss_simple}
y_{t} = \beta^{\TRANS} z_{t} + w_{t}, \quad t\geq 0,
\end{equation}
where $\beta \in \reals^{n}$ is an unknown parameter, $z_{t} \in \reals^{n}$ is $\mathcal{G}_{t-1}$-measurable covariate process, $y_{t}$ is the observation process, and  $w_{t} \in \reals$ is a noise process satisfying Assumption~\ref{ass:noise} with $\mathcal{F}_{t}$ replaced by $\mathcal{G}_{t}$. Then the least squares estimate  $\hat{\beta}_{T}$ of $\beta$ is given by:
\begin{equation}\label{eq:lai_regression}
\hat{\beta}_{T} = \argmin_{\beta^{\TRANS}}\sum_{\tau=0 }^{T}\|y_{\tau} - \beta^{\TRANS} z_{\tau}\|^2.
\end{equation}
The following result by \cite{lai1982least} characterizes the rate of convergence of $\hat{\beta}_{T}$ to $\beta$ in terms of unnormalized covariance matrix of covariates $Z_{T} \coloneqq \sum_{\tau=0}^{T}z_{\tau}z^{\TRANS}_{\tau}$.  
\begin{theorem}[{see \cite[Theorem~1]{lai1982least}}]\label{thm:lai_regression}
	Suppose the following conditions are satisfied: \textup{\textbf{(S1)}} $\lambda_{\min}(Z_{T}) \rightarrow \infty$, a.s. and  \textup{\textbf{(S2)}} $\log(\lambda_{\max}(Z_{T})) = o(\lambda_{\min}(Z_{T}))$, a.s. Then the least squares estimate in \eqref{eq:lai_regression} is strongly consistent with the rate of convergence:
	\[
	\|\hat{\beta}_{T} - \beta\|_{\infty} = \OO\bigg(\sqrt{\frac{\log\big[\lambda_{\max}(Z_{T})\big]}{\lambda_{\min}(Z_{T})}}\bigg) \quad \text{a.s.}
	\]
\end{theorem}
 Theorem~\ref{thm:lai_regression} is valid for all the $\mathcal{G}_{t-1}$-measurable covariate processes $\{z_{t}\}_{t \geq 0}$. For the switched least squares system identification if we take $\mathcal{G}_{t}$ to be equal to $\mathcal{F}_{t}$ and verify conditions (S1) and (S2) in Theorem~\ref{thm:lai_regression}, then we can use Theorem~\ref{thm:lai_regression} to establish its strong consistency and rate of convergence. As mentioned earlier in Remark~\ref{rem:covraiates}, the empirical covariances are coupled across different components due to the system dynamics. 
\subsection{Proof of Theorem~\ref{thm:main_1} } \label{subsec:proof_main_1}
To prove this theorem, we check the sufficient conditions in Theorem~\ref{thm:lai_regression}. First requirement that $X_{i,T}$ is measurable w.r.t. $\mathcal{F}_{T-1}$, follows by the definition of $X_{i,T}$. Conditions (S1) and (S2) are verified in the following.
\begin{enumerate}
	\item[(S1)] By Proposition~\ref{prop:main}-(P2), we see that $\lambda_{\min}(X_{i,T}) \rightarrow \infty$ a.s.; therefore, (S1) in  Theorem~\ref{thm:lai_regression} is satisfied.
	\item[(S2)]  Proposition~\ref{prop:main}-(P1) and (P2) imply that there exist positive constants $C_{1},C_{2}$, such that :
	\begin{multline*}
	\limsup_{T\rightarrow \infty}\frac{\log(\lambda_{\max}(X_{i,T}))}{\lambda_{\min}(X_{i,T})}  \notag\\ \leq \limsup_{T \rightarrow \infty} \frac{\log(C_{1}) + \log(T)}{C_{2}|\mathcal{T}_{i,T}|} = 0 \quad \text{a.s.}
	\end{multline*}
	where the last equality follows by Lemma~\ref{lem:occ_measure} (i.e. $|\T_{i,T}| = \OO(T)$, a.s.). Therefore, the second condition of Theorem~\ref{thm:lai_regression} is satisfied.
\end{enumerate}
Therefore, by Theorem~\ref{thm:lai_regression}, for each $i \in \mathcal{S}$, we have:
\begin{equation}\label{eq:pf_1}
 \bigl\| \hat A_{i,T} - A_i \bigr\|_\infty \le
\OO\bigg(\sqrt{\frac{\log \big[\lambda_{\max}(X_{i,T})\big]}{\lambda_{\min}(X_{i,T})}}\bigg), \quad \text{a.s.}   
\end{equation}
which proves the claim in Theorem~\ref{thm:main_1}.

\subsection{Proof of Corollary~\ref{cor:2}}\label{subsec:pf_cor2} 
Corollary~\ref{cor:2} is the direct consequence of Theorem~\ref{thm:main_1} and Proposition~\ref{prop:main}. Proposition~\ref{prop:main}-(P1) implies that $\lambda_{\max}(X_{i,T}) = \OO(\log(T))$. By substituting $\lambda_{\max}(X_{i,T})$ with $\OO(\log(T))$ in the right hand side of Eq.~\eqref{eq:pf_1}, we get that for each $i \in \mathcal{S}$, the estimation error $\| \hat A_{i,T} - A_i \bigr\|_\infty$ is upper-bounded by $\OO\big(\sqrt{\log(T)/|\mathcal{T}_{i,T}|}\big)$, $a.s$.


\subsection{Proof of Theorem~\ref{thm:main_2}}\label{subsec:thm_2} We first establish the strong consistency of the parameter $
\hat{\theta}_{T}$. By Theorem~\ref{thm:main_1} and the fact that $k < \infty$, we get:
\[
\bigl\| \hat{\theta}_{T} - \theta \bigr\|_\infty \le
\max_{i \in \mathcal{S}}\OO\bigg(\sqrt{\frac{\log \big[\lambda_{\max}(X_{i,T})\big]}{\lambda_{\min}(X_{i,T})}}\bigg), \quad \text{a.s.}
\]
Therefore, the result follows by applying Theorem~\ref{thm:main_1} to the argmax of above equation. For the second part notice that by Lemma~\ref{lem:occ_measure}, we know $\lim_{T \rightarrow \infty} |\T_{i,T}|/T = \pi_{\infty}(i)$, a.s.
Now, by Corollary~\ref{cor:2}, we get:
\[
\bigl\| \hat A_{i,T} - A_i \bigr\|_\infty \le
\OO\bigg(\sqrt{\frac{\log(T)}{|\mathcal{T}_{i,T}|}}\bigg) = 
\OO\bigg(\sqrt{\frac{ \log T }{\pi_{\infty}(i)T}}\bigg) \quad \text{a.s.}
\]
which is the claim of Theorem~\ref{thm:main_2}.

\section{Discussion on stability in the average sense}\label{sec:stab_avg}
The main results of this paper are derived under Assumption~\ref{ass:stab_avg} i.e., the MJS system~\eqref{eq:model} is stable in the average sense. In this section, we discuss the connection between this notion of stability  and more common forms of stability, i.e., mean square stability and almost sure stability.
\subsection{Stability on the average sense and mean square stability}		
	A common assumption on the stability of MJS systems (e.g., \cite{sarkar2019nonparametric} and \cite{sattar2020non}) is mean square stability defined as following:
\begin{definition}
	The MJS system~\eqref{eq:model} is called mean square stable (MSS) if there exists a deterministic vector $x_{\infty} \in \reals^{n}$ and a deterministic positive definite matrix $Q_{\infty}\in \reals^{n \times n}$ such that for any deterministic initial state $x_{0}$ and $s_{0}$ , we have:
	$
	\lim_{\tau \to \infty}\bigl\|\EXP[x_{\tau}] - x_{\infty} \bigr\| \to 0$, and $
	\lim_{\tau \to \infty}\bigl\|\EXP[x_{\tau}x^{\TRANS}_{\tau}] - Q_{\infty} \bigr\| \to 0.
	$
\end{definition}
\begin{proposition}[{see~\cite[Theorem~3.9]{costa2006discrete}}]
        The system is MSS, if and only if~
        $
        \lambda_{\max}\big((P^{\TRANS}\otimes I_{n^2})\diag(A_{i} \otimes A_{i})\big) <1.
        $
\end{proposition}

We now show that stability in the average sense is a weaker notion of stability than MSS.
\begin{proposition}\label{prop:mss-avg}
	If the MJS system~\eqref{eq:model} is mean square stable, then the system is stable in the average sense. 
\end{proposition}
The proof if presented in Appendix~\ref{sec:pf-mss-avg}.
\begin{remark}\label{rem:mss-avg}
	Proposition~\ref{prop:mss-avg} shows that MSS implies Assumption~\ref{ass:stab_avg}. Therefore, the results of Theorem~\ref{thm:main_1} and \ref{thm:main_2} also hold when Assumption~\ref{ass:stab_avg} is replaced by the assumption that the system is MSS.
\end{remark}

\subsection{Stability on the average sense and almost sure stability}
Consider the noise free version of the MJS system~\eqref{eq:model} with the following dynamics:
\begin{equation}\label{eq:noise_free}
	x_{t+1} = A_{s_t} x_t , 
	\quad t \ge 0.
\end{equation}
\begin{definition}
	The system~\eqref{eq:noise_free} is called almost surely stable if, for any deterministic initial state $x_{0}$ and $s_{0}$ we have:
	\[
	\lim_{t\to\infty}{\|x_{t}\|} = 0, \quad \text{a.s.}
	\]
	A common sufficient condition to check the almost sure stability of MJS system \eqref{eq:noise_free} is given below.
\end{definition}
\begin{proposition}[{see~\cite[Theorem~3.47]{costa2006discrete
}}] 
    If the stationary distribution $\pi_{\infty} = (\pi_{\infty}(1),\ldots,\pi_{\infty}(k))$  satisfies
	\textup{\textbf{(C1)}} $\pi_{\infty}(i)\not=0$ for all $i$ and \textup{\textbf{(C2)}} $\prod_{i=1}^k\sigma_{\max}(A_i)^{\pi_{\infty}(i)} < 1$, then the system~\eqref{eq:noise_free} is almost surely stable.
\end{proposition}
We now show that (C1) and (C2) are also sufficient conditions for stability in the average sense. 
\begin{proposition}\label{prop:cond_as}
	If the MJS system~\eqref{eq:model} satisfies (C1) and (C2), then the system is stable in the average sense. 
\end{proposition}
Proof is presented in Appendix~\ref{sec:pf_as}.
\begin{remark}
	Proposition~\ref{prop:cond_as} shows that (C1) and (C2) imply Assumption~\ref{ass:stab_avg}. Therefore, the results of Theorem~\ref{thm:main_1} and \ref{thm:main_2} also hold when Assumption~\ref{ass:stab_avg} is replaced by the assumption that the system satisfies (C1) and (C2).
\end{remark}
\subsection{Discussion on Non-Comparable  Stability Assumption}

The following examples illustrate that neither MSS nor conditions (C1) and (C2) in Proposition~\ref{prop:cond_as} is stronger than the other.

\begin{example}\label{ex:1}
	Let $\theta^{\TRANS}= \{A_{1}, 0\}$, and $p =(p_{1},p_{2})$ is an i.i.d. probability transition, with $\lambda_{\max}(p_{1}A_{1}) > 1$ and $x_{0} \not= 0$.  Then
	$
		\EXP[x_{\tau+1}] = \EXP [ A_{\sigma_{\tau}}x_{\tau} + w_{t}] =
		p_{1}A_{1}\EXP[x_{\tau}] = \cdots = (p_{1}A_{1})^{\tau} \EXP(x_{0})
	$
	, which implies
	$
	\lim_{\tau \rightarrow \infty} \EXP(x_{\tau}) = \infty.
	$	
	Therefore, this system is not mean square stable. However, this system satisfies conditions (C1) and (C2) in Prop.~\ref{prop:cond_as}  and therefore is stable in the average sense.  
\end{example}

\begin{example}\label{ex:2}
	Consider non-switched system with matrix $A$, with $\lambda_{\max}(A) < 1$ and $\sigma_{\max}(A) > 1$. 
	This system is mean square stable, but it doesn't satisfy the conditions (C1) and (C2) in Proposition~\ref{prop:cond_as}.
\end{example}

\section{Numerical Simulation}\label{sec:numerical_simulation}
In this section, we illustrate the result of Theorem~\ref{thm:main_1} via an
example. Consider a MJS with $n=2,k=2$, $A_{1}=\left[\begin{smallmatrix}
	1.5 &0\\
	0   &0.2
\end{smallmatrix}\right]$, and $A_{2}=\left[\begin{smallmatrix}
	0.01 &0.1\\
	0.1   &0.1
\end{smallmatrix}\right]$, probability transition matrix 
$
P = \left[\begin{smallmatrix}
	0.5 &0.5\\
	  0.75   &0.25
\end{smallmatrix}\right]$ 
 and i.i.d. $\{w_{t}\}_{t \geq 0}$  with $w_{t} \sim \mathcal{N}(0,I)$. Note that the example satisfies Assumptions~\ref{ass:noise} and conditions (C1) and (C2) of Proposition~\ref{prop:cond_as} (and, therefore, Assumption~\ref{ass:stab_avg}), but it is not mean square stable. We run the switched least squares for a horizon of $T = 10^{6}$ and repeat the experiment for $100$ independent runs. We plot the estimation error $e_{i,T} = \|\hat{A}_{i,t}-A_{1} \|_{\infty}$ versus time in Fig.~1. The plot shows that the estimation error is converging almost surely even though the system is not mean square stable. In Fig.~2,  logarithm of the estimation error versus logarithm of the horizon is plotted. The linearity of the graph along with approximate slope of $-0.5$ shows that $e_{i,T} = \tilde{O}(1/\sqrt{T})$.

\begin{figure}[ht]
	    \includegraphics[width=0.4\textwidth]{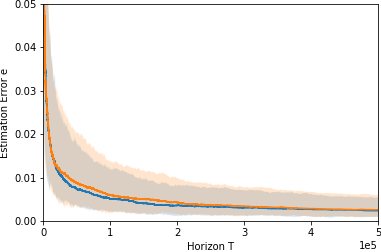}
	    \caption{Performance of switched least squares method for the example of Sec.~\ref{sec:numerical_simulation}. The solid line shows the mean across 100 runs and the shaded region shows the $25\%$ to $75\%$ quantile bound.}
\end{figure}
 
\begin{figure}[ht]
	\includegraphics[width=0.4\textwidth]{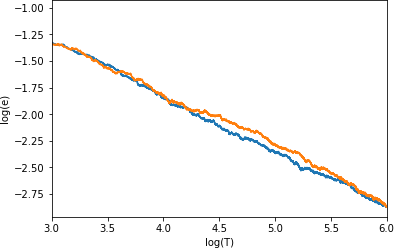}
	\caption{Logarithm of the estimation error versus logarithm of the horizon is plotted.   }
\end{figure}

\section{Conclusion and Future Directions}\label{sec:conclusion}
In this paper, we investigated system identification of (autonomous) Markov jump linear systems. We proposed the switched least squares method, showed it is strongly consistent and derived the almost sure rate of convergence of $\OO(\sqrt{\log(T)/T})$. This analysis provides a solid first step toward establishing almost sure regret bounds for adaptive control of MJS.

We derived our results assuming that system is stable in the average sense and we showed that this is a weaker assumption compared to mean square stability.  

The current results are established for autonomous systems with Markov switching when the complete state of the system is observed. Interesting future research directions include relaxing these modeling assumptions and considering controlled systems under partial state observability and unobserved jump times.

\bibliographystyle{IEEEtran}
\bibliography{IEEEabrv,mybibfile}
%
\appendices

\section{Proof of Proposition~\ref{prop:mss-avg}}\label{sec:pf-mss-avg}
\begin{proof}
		Since the system is MSS, there exists a positive definite matrix  $Q_{\infty} \in \reals^{n \times n}$ such that $\lim_{\tau \rightarrow \infty}\EXP[x_{\tau}x_{\tau}^{\TRANS}] = Q_{\infty}$, which implies $ \lim_{\tau \to \infty}\TR(\EXP[x_{\tau}x_{\tau}^{\TRANS}])=\TR(Q_{\infty})$. Since $\TR(\EXP[xx^{\TRANS}]) = \EXP[\TR(xx^{\TRANS})] = \EXP[x^{\TRANS} x]$, MSS implies that sequence of real numbers $\{\EXP(\|x_{\tau}\|^2)\}_{\tau \geq 0}$ converges to $\TR(Q_{\infty})$ and therefore:
		\begin{equation}\label{eq:pf-sum-E-finite}
			\lim_{T \rightarrow \infty}\frac{1}{T}\sum_{\tau=1}^{T}\EXP(\|x_{\tau} \|^2) = \TR(Q_{\infty}) < \infty
		\end{equation}
		Define events 
		\[
		E_{n} = \Big\{\omega \in \Omega :  \limsup_{T \rightarrow \infty}\frac{1}{T}\sum_{\tau=1}^{T}\|x_{\tau} \|^2 \leq n\Big\} ,\quad \forall n \in \integers
		\]
		and 
		\[
		E = \bigcup_{n =0}^{\infty}E_{n} = \Big\{\omega \in \Omega : \limsup_{T \rightarrow \infty}\frac{1}{T}\sum_{\tau=1}^{T}\|x_{\tau} \|^2 <\infty \Big\}.
		\]
		Now, by the continuity of probability measure from below, we have:
		\begin{equation}\label{eq:pf_cont_meas}
			\PR(E) = \PR(\bigcup_{n=0}^{\infty}E_{n}) =\lim_{n \to \infty} \PR(E_{n}).
		\end{equation}
		Note that 
		\begin{align*}
			\PR(E_{n}) & = 	\PR \big(\limsup_{T \to \infty}\frac{1}{T}\sum_{\tau=1}^{T}\|x_{\tau} \|^2 \leq n\big)\\ &\stackrel{(a)}{\geq} \limsup_{T \to \infty}\PR\big(\frac{1}{T}\sum_{\tau=1}^{T}\|x_{\tau}\|^2 \leq n\big)
			\\&\stackrel{(b)}{\geq} 1 -\limsup_{T \to \infty}\frac{\big(\sum_{\tau=1}^{T}\EXP\|x_{\tau}\|^2\big)}{Tn}
			\\&\stackrel{(c)}{\geq} 1-\frac{\TR(Q_{\infty})}{n},
		\end{align*}
		where $(a)$ follows from reverse Fatou's lemma, $(b)$ follows from the Markov  inequality and $(c)$ follows from Eq.~\eqref{eq:pf-sum-E-finite}. Substituting the above in equation (10), we get 
		\[
		\PR(E) \geq \lim_{n \to \infty} \Big(1-\frac{\TR(Q_{\infty})}{n}\Big) = 1.
		\]
		Therefore $\PR(E) = 1$, and the system is stable in the average sense. 
	\end{proof}

\vspace*{-9mm}
\section{Proof of Proposition~\ref{prop:cond_as}}\label{sec:pf_as}
\subsection{Asymptotic Behavior of Continuous Component }
To simplify the notation, we assume that $x_{0} = 0$  which does not entail any loss of generality. Let $\Phi(t-1,\tau+1) =A_{s_{t-1}}\cdots A_{s_{\tau +1}}$ denote the state transition matrix where we follow the convention that $\Phi(t,\tau) =I$, for $t < \tau$. Then we can write the dynamics in Eq.~\eqref{eq:model} of the continuous component of the state in convolutional form as:
\begin{equation}\label{eq:conv}
	x_{t} = \sum_{\tau=0}^{t-1} \Phi(t-1,\tau+1)w_{\tau}.
\end{equation}
where $\|\Phi(t-1,\tau+1)\| = \|A_{s_{t-1}}\ldots A_{s_{\tau+1}} \| $, and \begin{align}\label{eq:norm_bound}
    \|A_{s_{t-1}}\ldots A_{s_{\tau+1}} \| 
   \le \sigma_{s_{t-1}}\cdots\sigma_{s_{\tau+1}} &\eqqcolon  \Gamma_{t-1, \tau +1}
\end{align}
    where $\sigma_{s_{t}} = \sigma_{\max}(A_{s_{t}})$.
	In the following lemma, it is established that the conditions (C1) and (C2) in Prop.~\ref{prop:cond_as} imply that the sum of norms of the state-transition matrices are uniformly bounded. 
	\begin{lemma}[see {\cite[Lemma~1]{sayedana2021consistency}}]\label{lem:pre_lemma}
		Under the conditions (C1) and (C2) in Prop.~\ref{prop:cond_as}, there exists a constant $\bar{\Gamma} < \infty$ such that for all $T > 1$,
		$\sum_{\tau=0}^{T-1}\|\Phi(T-1,\tau+1) \| \leq \bar{\Gamma} $, $\text{a.s.}$
	\end{lemma}
The following Lemma shows the implication of Assumption~\ref{ass:noise} on the growth rate of energy of the noise process. 
\begin{lemma}[{\cite[Eq.~(3.1)]{lai1985asymptotic}}]
	Under Assumption~\ref{ass:noise} 
	$
		\sum_{\tau=0}^{T}\|w_{\tau} \|^2 = \OO(T), \quad \text{a.s.}
	$
\end{lemma}

Using the convolution formula in Eq.~\eqref{eq:conv}, we can bound the norm of the state $\|x_{t}\|^2$ as following:
\begin{align}\label{eq:x_bound}
	\|x_{t}\|^2  
	= &\Big(\big\|\sum_{\tau=0}^{t-1}\Phi(t-1,\tau+1)w(\tau)\big\|\Big)^2 \notag \\\stackrel{(a)}{\leq}\notag  &\Big(\sum_{\tau=0}^{t-1}\|\Phi(t-1,\tau+1)w(\tau) \|\Big)^2\nonumber\\\stackrel{(b)}{\leq}  &\Big(\sum_{\tau=0}^{t-1}\|\Phi(t-1,\tau+1) \| \|w(\tau)\|\Big)^2 \notag \\\stackrel{(c)}{\leq}  &\Big(\sum_{\tau=0}^{t-1}\Gamma_{t,\tau+1} \|w(\tau)\|\Big)^2
\end{align}
where $(a)$ follows from triangle inequality and $(b)$ follow from sub-multiplicative property of the matrix norm, and $(c)$ follows from Eq.~\eqref{eq:norm_bound}. Now for a fixed $i$, $i \in \mathcal{S}$, we have:

	\begin{align*}
		\sum_{t=0}^{T}\|x_{t} \|^2 &\stackrel{}{\leq} \sum_{t=0}^{T}\Big(\sum_{j=0}^{t-1} \Gamma_{j+1,t-1} \|w(j)\|\Big)^2\\&\stackrel{(d)}{\leq} \sum_{t=0 }^{T}\Big(\sum_{j=0}^{t-1}\Gamma_{j+1,t-1}\Big)\Big(\sum_{j=0}^{t-1}\Gamma_{j+1,t-1}\|w(j)\|^2\Big)\\
		&\stackrel{(e)}{\leq} \bar{\Gamma}\sum_{t =0}^{T}\Big(\sum_{j=0}^{t-1}\Gamma_{j+1,t-1}\|w(j)\|^2\Big) \\&\stackrel{(f)}{\leq} 
		\bar{\Gamma}\sum_{j=0}^{T-1}\Big(\sum_{t=0}^{T}\Gamma_{j+1,t-1}\Big)\|w(j)\|^2 \\&\stackrel{(g)}{\leq} \bar{\Gamma}^2\sum_{j=0}^{T-1}\|w(j) \|^2  = \OO(T) \quad \text{a.s.}
	\end{align*}
where $(d)$ follows from Cauchy-Schwarz's inequality, $(e)$ follows from Lemma~\ref{lem:pre_lemma}, $(f)$ follows from changing the order of summation, and $(g)$ follows from boundedness of sub-sums of $\sum_{\tau=0}^{T-1}\Gamma_{\tau+1,T-1} $, and Lemma~\ref{lem:pre_lemma}. 

\end{document}